\def\year{2020}\relax
\documentclass[letterpaper]{article} 
\usepackage{aaai20}  
\usepackage{times}  
\usepackage{helvet} 
\usepackage{courier}  
\usepackage[hyphens]{url}  
\usepackage{graphicx}  
\usepackage{amssymb}
\usepackage{amsmath}
\title{Towards Ghost-free Shadow Removal via \\
Dual Hierarchical Aggregation Network and Shadow Matting GAN }
\author{
Xiaodong Cun,\textsuperscript{\rm 1} 
Chi-Man Pun,\textsuperscript{\rm 1}\thanks{Corresponding Author} 
Cheng Shi\textsuperscript{\rm 1,2}\\ 
\textsuperscript{\rm 1} Department of Computer and Information Science, University of Macau, Macau, China\\ 
\textsuperscript{\rm 2} School of Computer Science, Xi'an University of Technology, Xi'an, China\\
\{yb87432,cmpun\}@umac.mo,chengc\_s@163.com 
}
\begin{document}

\maketitle

\begin{abstract}

Shadow removal is an essential task for scene understanding. Many studies consider only matching the image contents, which often causes two types of ghosts: color in-consistencies in shadow regions or artifacts on shadow boundaries~(as shown in Figure.~\ref{fig:detail}). In this paper, we tackle these issues in two ways. First, to carefully learn the border artifacts-free image, we propose a novel network structure named the dual hierarchically aggregation network~(DHAN). It contains a series of growth dilated convolutions as the backbone without any down-samplings, and we hierarchically aggregate multi-context features for attention and prediction, respectively. Second, we argue that training on a limited dataset restricts the textural understanding of the network, which leads to the shadow region color in-consistencies. Currently, the largest dataset contains 2k+ shadow/shadow-free image pairs. However, it has only 0.1k+ unique scenes since many samples share exactly the same background with different shadow positions. Thus, we design a shadow matting generative adversarial network~(SMGAN) to synthesize realistic shadow mattings from a given shadow mask and shadow-free image. With the help of novel masks or scenes, we enhance the current datasets using synthesized shadow images. Experiments show that our DHAN can erase the shadows and produce high-quality ghost-free images. After training on the synthesized and real datasets, our network outperforms other state-of-the-art methods by a large margin. The code is available: http://github.com/vinthony/ghost-free-shadow-removal/

\end{abstract}

\section{Introduction}
\noindent Shadows are a common phenomenon in nature. Cast shadows, which often appear in the opposite direction of a light source, form when the light from an illuminant is blocked by an object.  
Removing these shadows is essential because shadows may degrade the performance of many computer vision tasks, such as object detection and tracking~\cite{mikic2000moving,khan2015automatic}. Early approaches try to automatically detect shadow regions~\cite{Guo:2013tt,khan2015automatic} or to do so with user input~\cite{gong2014interactive,yang2012shadow} first, after which they transform the shadows to match the background. For deep learning-based approaches, \citeauthor{qu2017deshadownet}~\shortcite{qu2017deshadownet} train a neural network considering multi-context features~(global semantics and low-level details). Then, \citeauthor{wang2018stacked}~\shortcite{wang2018stacked} conduct joint shadow removal/detection using a stacked generative adversarial network. \citeauthor{Hu:2018wd}~\shortcite{Hu:2018wd} retrain their shadow detection network to directly train shadow removal network. More recently, \citeauthor{Anonymous:XfCvZjhb}~\shortcite{Anonymous:XfCvZjhb} train a network using an unpaired dataset without paired supervisions. 

\begin{figure}[t]
 \centering
  \includegraphics[width=\columnwidth]{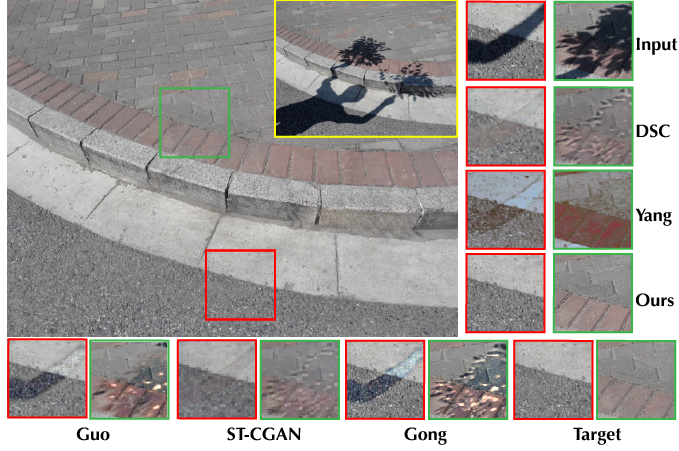}
  \caption{We plot a result of our model with the input shown in the yellow square. From the two zoomed regions, it can be seen that our method successfully removes the shadow and reduces the ghost.}
  \label{fig:detail}
\end{figure}

However, current deep learning-based techniques have two major drawbacks. First, we argue that previous network structures are not carefully designed for shadow removal. While the goal of shadow removal is restoring the color from the shadow image, the shadow border also plays a vital role in the visual quality. For example, DeShadowNet~\cite{qu2017deshadownet} learns only the shadow matting from multi-context features other than the full image using the pre-trained VGG19, and the artifacts will still show on the border. DSC~\cite{Hu:2018wd} learns using a direction-aware attention model, which well captures the directional details well and also preserves more shadow boundaries. Thus, a carefully designed model is criterial. In addition, the community lacks a high-quality image pairs~(shadow/shadow-free) for shadow removal and joint learning on shadow removal and detection~(shadow/shadow-free/mask). Since the largest available dataset contains only 0.1k+ unique scenes, and the performance of unsupervised learning is limited. We argue that such limited scenes will hugely influence the color consistency between the shadow and shadow-free regions because training on small dataset can not understand the scenes well.

In this work, we design a novel network structure to erase shadows and produce high-quality shadow-free images according to the following observations: \textit{the shadow and shadow-free images share the same semantic information and shadow removal specifically and only needs to learn the shadow}. Since the semantic knowledge is the same and the network needs to focus on low-level features, different from others, an image processing network~\cite{chen2017fast} serves as a useful starting point for our method. In detail, this network uses several dilated convolutions~\cite{yu2015multi} linearly with a growing dilated rate to capture multi-contexts and preserve useful low-level features. Based on this backbone, we hierarchically aggregate dilated convolutions to achieve better feature reuse~\cite{Yu:2017ts}. Then, to specifically learn the shadow regions, we design the hierarchical aggregation attention model with the help of multi-contexts and the attention loss from the shadow mask. Non-linear feature aggregation helps our attention to gain the knowledge from multiple previous layers, and dilated convolutions preserve the details well.

In addition, we get our inspiration from the nature in that a cast shadow comes from the opposite direction of the light source. When an object moves, a single shadow-free image can produce countless shadow images. Thus, if we train the network to synthesize shadow images from the corresponding masks and shadow-free images, the network will also produce the novel shadows according to the novel mask. It is much easier to collect the required data for shadow synthesis than for triples of paired shadow/shadow-free/mask. Thus, we design a novel shadow synthesis network using GAN~\cite{goodfellow2014generative} to enlarge the current datasets. Instead of learning the shadow image directly, our network learns the shadow matting first and produces the high-quality shadow from the original shadow-free image and matting.
Although these images are synthesized using neural network, they still have a similar distributions to the real natural scene compared with those from computer games~\cite{sidorov2019conditional}. In our shadow synthesis, we assume the scenes contain only cast shadows with the objects outside the scene. Experiments shows that our setting also works well in other situations.

 We summarize our major contributions as follows.
\begin{itemize}

\item We design a joint learning framework, named Dual Hierarchical Aggregation Network~(DHAN), for shadow removal with the help of shadow detection. The proposed network hierarchically aggregates the dilated multi-contexts features and attentions, respectively.
\item We design and train a Shadow Matting GAN on current paired shadow datasets. Then, the novel paired samples are obtained from the well-trained network with novel masks or scenes. Finally, we train shadow-related tasks using augmented datasets.
\item Experiments on real and synthetic data indicate that our algorithm outperforms other state-of-the-art methods, especially on the visual quality. 
\end{itemize}

\section{Related Work}

\textbf{Shadow Removal}
Early works often erase shadows via user interaction or hand-crafted features~\cite{Guo:2013tt,khan2015automatic}. Inspired by the convolution neural network-based method, \citeauthor{qu2017deshadownet}~\shortcite{qu2017deshadownet} propose a multi-stream neural network for removing the shadows according to global-and-local contexts, and \citeauthor{Hu:2018wd}~\shortcite{Hu:2018wd} use a fully convolutional network structure but with direction-aware attention. For learning shadow removal and detection jointly, \citeauthor{wang2018stacked}~\shortcite{wang2018stacked} proposed a stacked GAN where the learned shadow mask serves as the attention for the removal. \citeauthor{BinDing:2019wa}~\shortcite{BinDing:2019wa} extend this method by strong network and learning recurrently.

\noindent\textbf{Shadow Synthesis}
For it is hard to collect larger shadow removal datasets, 
\citeauthor{Gryka2015softShadows}~\shortcite{Gryka2015softShadows} generate the shadow images from graphics tools and \citeauthor{sidorov2019conditional}~\shortcite{sidorov2019conditional}  generate the shadow dataset by controlling the lights in the video game. However, scenes or shadows that are rendered by computer graphics are significantly different from natural scenes. \cite{Le:2018vu,Le-etal-ICCV19} propose a two-stage network for generating the adversarial shadow images before shadow detection, and they also use this idea to remove shadows by enhancing the shadow strength. Focusing on the limitation of a paired dataset, \citeauthor{Anonymous:XfCvZjhb}~\shortcite{Anonymous:XfCvZjhb} train a neural network to remove shadows from unpaired shadow/shadow-free images. It also contains an auxiliary network for shadow synthesis. However, their synthesized shadows are not learned using the paired dataset and their shadow generator is used in cycle-consistency loss only.

\noindent\textbf{Shadow Detection}
Some traditional methods use physical models of illumination and color for shadow detection~\cite{salvador2004cast,panagopoulos2011illumination}. Others describe image regions by hand-crafted features, such as color~\cite{lalonde2010detecting}, texture~\cite{vicente2017leave}, then recognize these regions by classifier. However, these approaches are not well suited for various natural scenes.
Current state-of-the-art shadow detection methods are based on the power of convolution neural network, such as learning shadow edges~\cite{khan2015automatic,shen2015shadow} and 
designing attention modules by considering the direction~\cite{Hu:2018wd}, recurrent attention~\cite{zhu2018bidirectional} and distraction~\cite{zheng2019distraction}.

\begin{figure*}[t]
	\centering
  \includegraphics[width=0.9\textwidth]{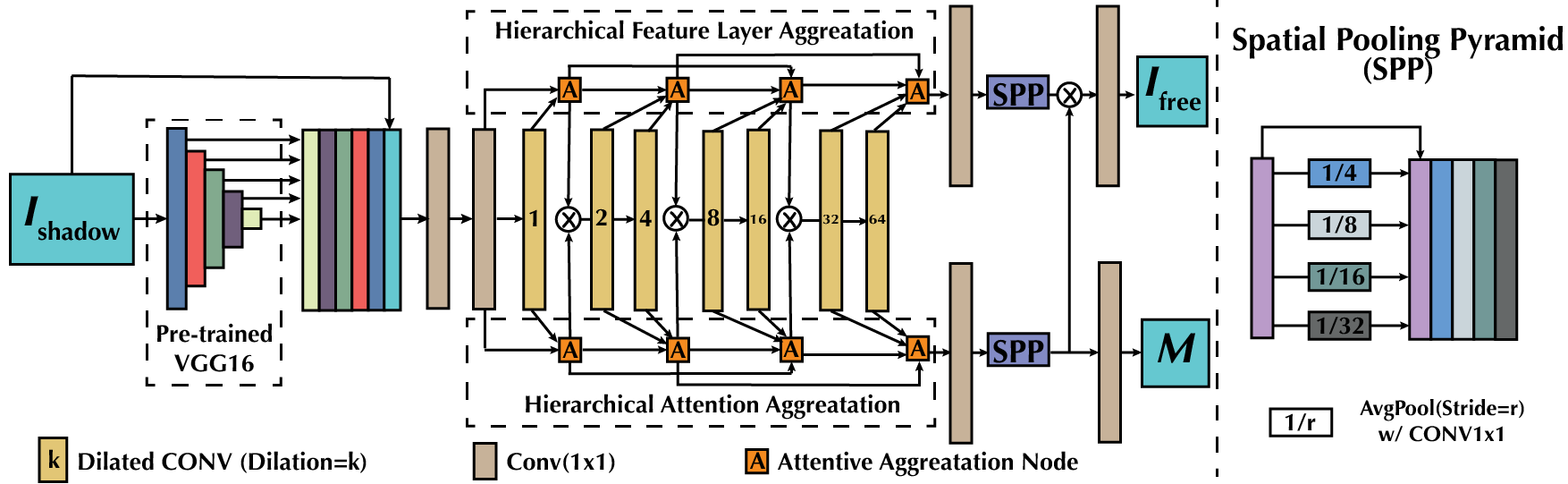}
  \caption{The network structure of the proposed  Dual Hierarchical Aggregation Network.}
  \label{fig:network_structure}
\end{figure*}

\begin{figure}[b]
\centering
\includegraphics[width=0.8\columnwidth]{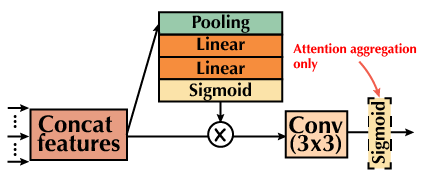}
  \caption{Attentive Aggregation nodes for feature aggregation and attention aggregation.}
  \label{fig:aan}
\end{figure}

\section{Proposed Methods}

\subsection{Dual Hierarchical Aggregation Network}

Given a shadow image $I_{shadow}$, our method learns the shadow-free image $I_{free}$ using the proposed network with the help of the attention loss~(shadow mask $M$) in a single forward process. 
As previously discussed, our backbone structure is based on the context aggregation network~(CAN)~\cite{chen2017fast,2018arXiv180605376Z} for image processing. Specifically, we extract the hyper-column features from the pre-trained VGG19~(CONV1\_2, CONV2\_2, CONV3\_2, CONV4\_2 and CONV5\_2) and the original RGB image serves as an augmented network input. Then, we encode these multi-context features using several dilated convolutions with a growing dilate rate. Each convolutional block follows an Instance Normalization and LeakyReLU layer. Previous studies show that this network structure works well in image processing tasks, such as Nonlocal de-hazing and $L_0$ smoothing since there is no down-sampling or stride convolution in the context encoding, the low-level details can be well preserved. Although shadow removal is similar to low-level task, it specifically needs to learn the shadow regions. Possible solutions are to learn the particular region using partial convolution~\cite{liu2018image} or gated convolution~\cite{yu2018free}.  However, these attention mechanisms are designed for image in-painting. Thus, they need to fill the region using global semantic contexts and work in a Unet-like structure. 

Thus, to specifically learn the shadows and preserve low-level details, we design \textbf{dual hierarchical aggregations} for spatial attentions and mixed layer features, respectively. In detail, as shown in Figure.~\ref{fig:network_structure}, we build the attention module by aggregating the features from multiple previous layers in a hierarchical layer aggregation style~\cite{Yu:2017ts}. Our $n$ layer dual hierarchical aggregation network $DHAN_{n} = T_{n} \times AT_{n}$ merges the features in a tree-like structure for richer context attention. In it, $T_{n}$ and $AT_{n}$ can be defined as:
\begin{equation}
\begin{aligned}
T_{n} = N(R^n_{n-1}(x),...,R^n_{1}(x),
L^n_1(x),L^n_2(x))\\
AT_{n} = AN(AR^n_{n-1}(x),...,AR^n_{1}(x),L^n_1(x),L^n_2(x))
\end{aligned}
\end{equation}
where $N$ and $AN$ are the aggregation node and attention aggregation node, respectively. $N(\cdot) $ means the feature concatenation. $R$, $AR$ and feature layer $L$ are defined as:
\begin{equation}
\begin{aligned}
L^n_1(x) = DB_{k=2^{n-1}}(R^n_1(x)\times AR^n_1(x))\\
L^n_2(x) = DB_{k=2^{n}}(L^n_1(x))\\
R^n_m(x)=
\begin{cases}
T_m& \text{if m = n-1}\\
T_m(R^n_{m+1}(x))& \text{otherwise,}
\end{cases}
\\
AR^n_m(x)=
\begin{cases}
AT_m& \text{if m = n-1}\\
AT_m(AR^n_{m+1}(x))& \text{otherwise,}
\end{cases}
\end{aligned}
\end{equation}
where $DB_{k=x}$ represents the $x$-dilated convolution.  
Thus, each aggregated feature is spatially reweighed by the multi-context attention. In addition, we also use the ground truth shadow mask as the supervision for the attention. 

In each aggregation node~(as shown in Figure.~\ref{fig:aan}), we use a squeeze-and-excitation block~\cite{hu2018squeeze} to re-weight the importance of each feature channel. Then, a $3\times3$ convolution is added to squeeze the features and match the original channels. Notice that, the $Sigmoid(\cdot)$ layer is added only at the end of the attention aggregation nodes~($AR$). Finally, a spatial pooling pyramid~(as shown in Figure.~\ref{fig:network_structure})~\cite{he2015spatial} is added at the end of the last aggregation block for multi-context features remixing.

\begin{figure*}[h]
	\centering
  \includegraphics[width=\textwidth]{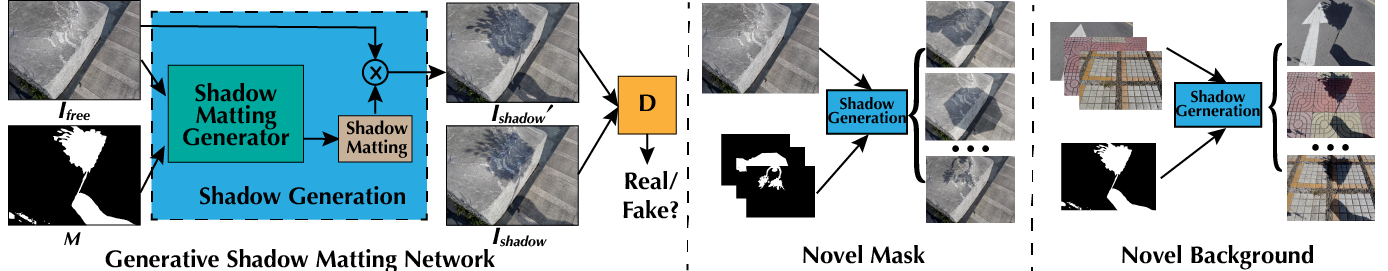}
  \caption{The network structure of Shadow Matting GAN.}
  \label{fig:shadowbank}
\end{figure*}

\subsubsection{Loss Functions}
By considering the semantic measures and low-level details in multiple contexts, we train our network using the multi layer perception loss~\cite{2018arXiv180605376Z,johnson2016perceptual} $L_{\Phi}$ in the pixel and feature domain as follows: 
\begin{equation}
  L_{\Phi} = \sum^{5}_{k=0}\lambda_{l}||\Phi_{k}( I_{free}^{'})) - \Phi_{k}(I_{free}) ||_1
 \label{eq:vgg} 
\end{equation}
where ${\Phi}$ is the VGG16 that is pre-trained on ImageNet. We estimate the differences from the $\rm{CONV} k\_2 (k = 1...5) $ and the original images~($k=0$ in Equation.~\ref{eq:vgg}) between the natural shadow-free image $I_{free}$ and the generated $I_{free}^{'}$ using the $L_1$ loss with weight $\lambda_{l}$. For the attention loss, we use the binary cross entropy~(BCE) loss between the predicted mask $M'$ and the ground truth mask $M$ as follows:
\begin{equation}
\begin{aligned}
L_{m} = \frac{1}{N}\sum- [Mlog(M') + (1- M)log(1-M')]  \\
 \end{aligned}
\end{equation}
Where $N$ is the total number of pixels.
In addition, the convolution neural network might also preserve the shadow border in the images, which will also degrades the shadow removal quality. Thus, we use the adversarial loss as \cite{isola2017image} in our approach. By considering our DHAN as the generator, the discriminator $D$ is constructed with several convolution blocks. We simply discriminate the patches in the real shadow-free image $I_{free}$ and the predicted $I_{free}^{'}$ given the shadow image $I_{shadow}$. The discriminator is optimized to identify the generated image that pushes the generator towards realistic shadow-free images. The discriminator loss is as follows:
\begin{equation}
\begin{aligned}
L_{cGAN} = \sum [log(D(I_{shadow},G(I_{shadow})) \\
- log(D(I_{shadow},I_{free})) ]
\end{aligned}
\label{eq:gan} 
\end{equation}
Overall ,the final objective of our method $G^{*}$ is as follows:
$\mathop{\arg\min}_{G}\mathop{\arg\max}_{D}L_{cGAN} + \lambda L_{\Phi} + \alpha L_{m}$. We set $\alpha$ equals to 100 and $\beta$ equals to 20 empirically.

\subsection{SM-GAN: Shadow Matting GAN}
 As discussed in the introduction, one of the key issues in current shadow removal methods is the limitation of the paired dataset. However, with respect to the various types of shadow masks and scenes, it is hard to collect a large dataset with paired shadow/shadow-free images. The state-of-the-art SRD dataset~\cite{qu2017deshadownet} and ISTD~\cite{wang2018stacked} contains only 100+ different scenes and 10+ unique shadow types. In addition,
the environmental light changes so fast that it is impossible to take pairs of shadow/shadow-free images under the same light conditions, even in a flash. The uncertainty of the light transforming between shadow/shadow-free images influences the accuracy of the network. For these reasons, we argue that the application range of current learning-based shadow removal algorithms is limited and training using limited scenes will cause color inconsistencies between shadow and shadow-free regions. However, it is far easier to collect the natural shadow-free images and random masks than to collect the corresponding shadow/shadow-free/shadow-mask sets. Thus, we synthesize the shadow image using the random shadow mask and the shadow-free image via the GAN to improve the accuracy of shadow networks and extend their capabilities to larger natural domains. 

\begin{figure*}[t]
\centering     
\includegraphics[width=\textwidth]{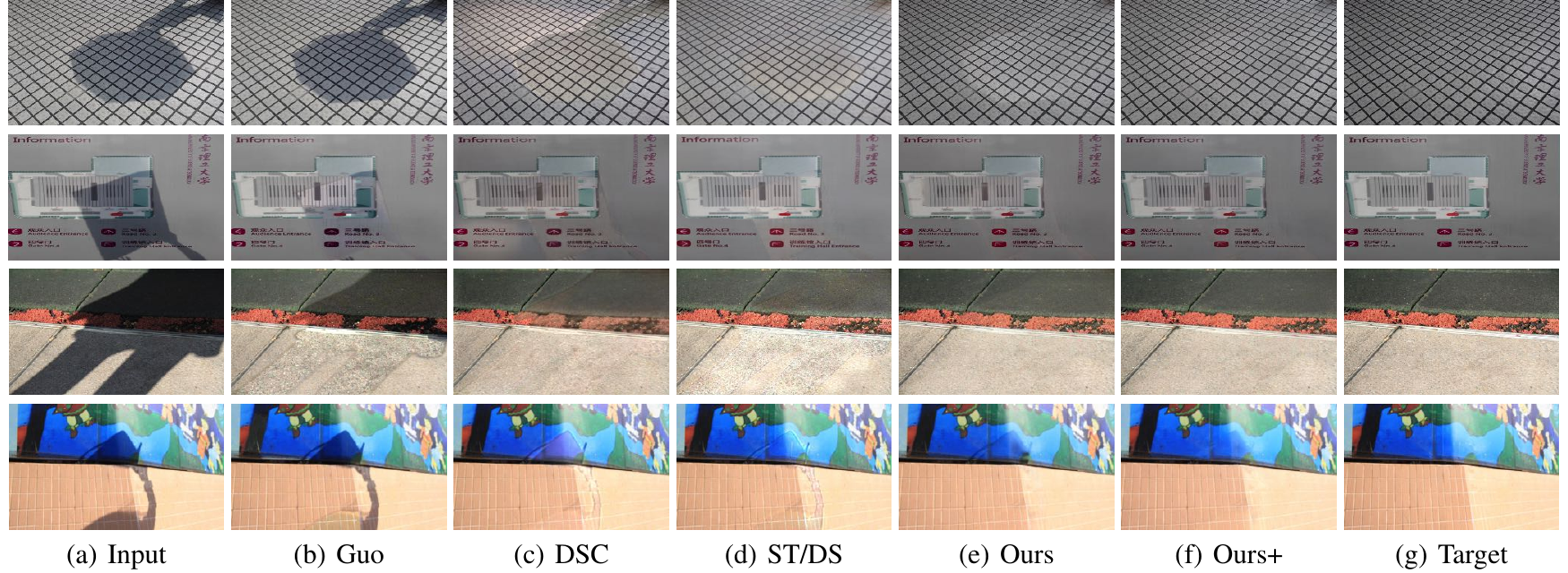}
\caption{Shadow Removal Comparison. The top and bottom two samples are from ISTD and SRD dataset, respectively. In (d), the top and bottom two results are from ST-CGAN~(ST) and DeShadowNet~(DS), respectively. Best view with zoom-in.}
\label{fig:srd}
\end{figure*}

\begin{table*}[t]
\begin{center}
\caption{Comparison of Shadow removal results. The subscripts represent the years of the compared methods.}
\label{table:rmse}
\begin{tabular}{|l|c|c|c|c|c|c|c|c|c|c|}
\hline
Method & S & NS & RMSE  & SSIM-S  & PSNR-S & S & NS & RMSE  & SSIM-S  & PSNR-S  \\
\hline
& \multicolumn{5}{|c|}{ISTD dataset} & \multicolumn{5}{|c|}{SRD dataset} \\
\hline
Original  & 32.67 & 6.83 & 10.97 & 98.19 & 32.87 & 46.87 & 5.60 & 14.28 & 88.54 & 19.35 \\
Yang$_{12'}$  &  19.82      &  14.83     & 15.63   & 93.25 & 28.01  &  23.43      &  22.26     & 22.57  & -  &  - \\
Guo$_{13'}$  &  18.95      &   7.46     & 9.30   & 96.04 & 27.49 &  29.89     &   6.47     & 12.60   & 91.46 & 23.69   \\
Gong$_{14'}$  & 14.98 &  7.29 & 8.53  & 96.97 & 29.79 & 19.58 &  4.92 & 8.73  & - & - \\
DeShadowNet$_{17'}$  &  -   & - & - & - &  - &  10.81 & 4.85 & 6.10  & 94.68 & 31.97 \\
ST-CGAN$_{18'}$    &  10.33   & 6.93 & 7.47  & 94.93  &  32.23 & - & - & -&- & -\\
DSC$_{18'}$  &   9.48      &  6.14   & 6.67   &   96.66     &  33.45  & 10.89  & 4.99 &  6.23  & 93.75  & 31.69 \\
Ours~(DHAN)  &  \underline{8.14} & \underline{6.04} & \underline{6.37} & \underline{98.29} & \underline{34.50} &  \underline{8.94} & \underline{4.80} & \underline{5.67} & \underline{95.29} & \underline{33.36} \\
Ours~(DHAN)+DA    & \textbf{7.52}  & \textbf{5.43} & \textbf{5.76} & \textbf{98.36} & \textbf{34.98}  & \textbf{8.39} & \textbf{4.67} & \textbf{5.46} & \textbf{95.31}  & \textbf{33.72} \\
\hline
\end{tabular}
\end{center}
\end{table*}

As shown in Figure.\ref{fig:shadowbank}, using the knowledge of the shadow image~$I_{shadow}$, the shadow mask~$M$ and the corresponding shadow-free image~$I_{free}$ in the current dataset, we train shadow synthesis as a paired task~\cite{isola2017image} using the GAN. Specifically, we train a generator to synthesize the shadow image $I_{shadow}^{'}$ from the corresponding shadow-free image $I_{free}$ and mask $M$. Then, the discriminator is optimized to identify whether the synthesized shadow image $I_{shadow}^{'}$ is real or not. Since there is a corresponding ground truth image $I_{shadow}$ in the dataset, apart from adversarial loss from the discriminator, we restrict the similarity of the $I_{shadow}^{'}$ and $I_{shadow}$ in a supervised manner. Instead of generating the shadow image directly, we learn the shadow matting $I_{matting}$ following the former observation~\cite{qu2017deshadownet}: $I_{shadow} =  I_{matting} \times I_{free}  $. Thus, our shadow synthesis network can be written as follows:
\begin{equation}
  I_{shadow}^{'} = G(I_{free},M) \times I_{free}
\end{equation}
where $G(\cdot)$ is the generator of our shadow matting GAN. The supervised losses in the shadow matting generator are similar to our shadow removal: We use the multi-scale perceptual loss to measure the prediction quality and adversarial loss to guarantee the realism of the output.
Then, we update the parameters in the discriminator network $D$ to recognize whether the generated shadow image is real or not. The generator of our SM-GAN has a similar structure with the CycleGAN~\cite{Zhu:2017ua}. 
For simplicity, the discriminator is constructed with 5 convolution layers with ReLU non-linear activation and Batch Normalization layer, which is  similar to our shadow removal network.

After training, the generator in SMGAN can be easily used to sample novel domain examples, such as the novel background and novel shadow mask~(as shown in the novel mask/background of Figure.\ref{fig:shadowbank}). 
Here, we use the shadow-free images in USR~\cite{Anonymous:XfCvZjhb} and shadow masks from ISTD training set to synthesize the shadow images for data augmentation. For each shadow-free image in the USR dataset, we randomly choose 3 images from the training mask of the ISTD dataset to generate shadow. The experiments show that this kind of setting is sufficient for the shadow removal and detection tasks. 

\section{Experiments}
In this section, first, we conduct the shadow removal experiments using  standard benchmarks, the ISTD~\cite{wang2018stacked} and the SRD~\cite{qu2017deshadownet}. The SRD contains 2k+ and 408 images for training and testing, respectively. In ISTD, there is 1.3k and 540 images  for training and testing, respectively. On both datasets\footnote{Since there is no ground truth shadow mask in the SRD, we calculate the shadow matting using the shadow/shadow-free pairs and generate the binary mask via thresh-holding.}, we jointly learn the shadow removal and attention~(shadow mask) using the proposed DHAN. In addition, we retrain these tasks on the dataset containing both real and synthesized shadow images. Then, we evaluate the effectiveness of our synthesized shadows for shadow detection on SBU~\cite{vicente2016large} dataset, which contains 4k+ and 0.6k+ images for training and testing, respectively. Finally, we conduct the ablution study on the network structure, the quality of shadow synthesis and the network capabilities.

\subsubsection{Implementation Detail}
We train all our models using TensorFlow framework and optimize the network on Adam optimizer \cite{kingma2014adam} with a fixed learning rates of $2*10^{-4}$ and $10^{-4}$ in the generator and the discriminator, respectively. In training, we randomly resize the original image from 256p to 480p, thereby retaining the original aspect radio rather than cropping the image. We set the batch size to 1 and train the network in over 100 epochs for shadow synthesis and 150 epochs for shadow removal~(60 epochs for the augmented datasets). For testing, we evaluate all the images in their original size. It takes 0.2s for our network to generate a shadow-free image with a resolution of 640x480 on a single NVIDIA TITAN V GPU. On shadow removal, we follow the previous works and evaluate the RMSE on the LAB color space in the shadow~(S) and non-shadow~(NS) regions. In addition, we also report the PSNR and SSIM for the shadow region~(SSIM-S and PSNR-S, respectively) to evaluate whether it can produce high-quality shadow removal results. To evaluate on shadow detection, we use the Balance Error Rate~(BER) for the shadow~(S) and non-shadow region~(NS) as in previous works. The BER is defined as follows: $BER = 1 - \frac{1}{2}(\frac{TP}{TP+FN}+\frac{TN}{TN+FP})$, 
where TP, TN, FP and FN are the total numbers of the true positives, true negatives, false positives and false negatives, respectively.

\begin{figure*}[t]
\centering     
\includegraphics[width=\textwidth]{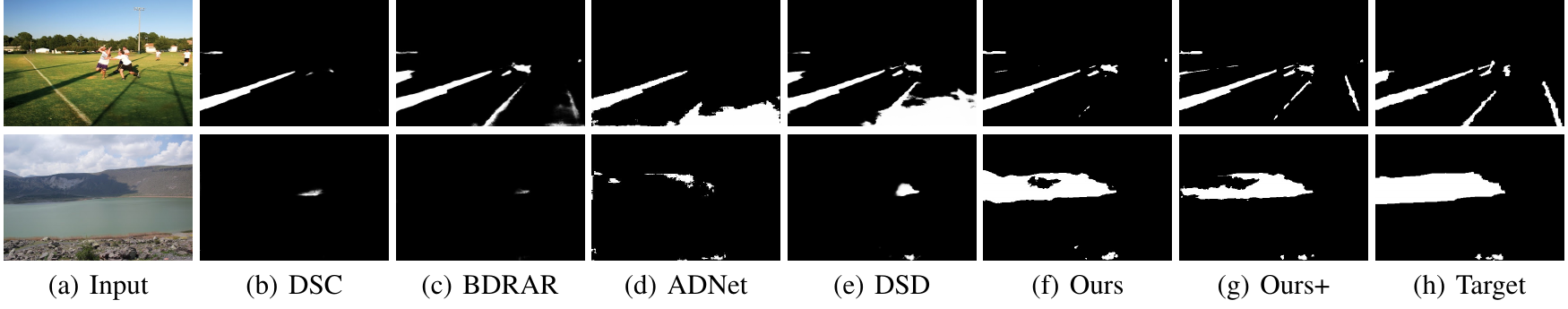}
\caption{Comparison the shadow detection results with state-of-the-art methods on the SBU dataset.}
\label{fig:sbu}
\end{figure*}

\subsection{Comparison with the State-of-the-art Methods}
\subsubsection{Shadow Removal}
We compare ours results with those of the state-of-the-art shadow removal methods, including ST-CGAN~\cite{wang2018stacked} that jointly conducts shadow removal and detection, DSC~\cite{Hu:2018wd} that removes shadows using a direction-aware attention mechanism, and DeShadowNet~\cite{qu2017deshadownet} conducts shadow removal using multi-context features and the shadow matting loss. Apart from deep learning-based methods, we also compare our algorithm with some open-sourced traditional methods~\cite{Guo:2013tt,gong2014interactive,yang2012shadow}. Note that, in order to conduct a fair comparison, all the results and figures are provided by the authors or taken from the original papers. Since the training codes of ST-CGAN and DeShadowNet are not publicly available, we evaluate their performances only on their own dataset. As shown in Figure.~\ref{fig:srd}, our method provides much better visual quality than the others on both two standard datasets. In these samples, previous works produce many ghosts because of the remaining shadow boundaries or color inconsistencies. For example, DeShadowNet learns the shadow matting only where the restored color might not match the background. In addition, the DSC and DeShadowNet will cause sharp shadow boundaries since their network structures need to down-sample and up-sample in the multi-contexts. Meanwhile, due to the limited dataset, they all lack better scene understandings. However, the proposed method~(Ours) has minimal artifacts in the shadow boundaries and the synthesized dataset used as augmentation~(Ours+DA) provides much better color consistency. For the numerical evaluation in Table.~\ref{table:rmse}, all the methods can successfully learn the shadow-free region with the help of network; however, it is clear that our DHAN outperforms the others, especially on the shadow regions, because of the effectiveness of our attention model and attention loss. Additionally, training on the augmented synthesized shadow dataset helps a lot in shadow and non-shadow regions since it successfully enlarges the knowledge of current dataset. Note that, on SRD dataset,
even though ISTD dataset contains larger color differences between the training pairs in the shadow-free region and our network try to fit these mapping in shadow synthesis, the synthesized shadows also help the scenes understanding on SRD dataset. 

\subsubsection{Shadow Synthesis for Detection} Our synthesized dataset also benefits to shadow detection. In detail, we modify our network structure to match the detection by removing the output of attention loss. Thus, we only use the mask as output for back-propagation. We compare our results with state-of-the-art learning-based shadow detection methods, such as attention-based methods DSC~\cite{Hu:2018wd}, BDRAR~\cite{zhu2018bidirectional}, DSD~\cite{zheng2019distraction}, joint learning method ST-CGAN~\cite{wang2018stacked} and adversarial shadow-based methods ADNet~\cite{Le:2018vu}. As shown in Table.~\ref{table:detection}, although our network is not particularly designed for shadow detection, it still gets better results than those other methods under the same backbone. Our network also obtains comparable results with the state-of-the-art methods using the powerful pre-trained ResNext101 model and multi-contexts supervisions. The shadows in the SBU dataset are much more challenging and complex than simple cast shadows. However, our synthesized shadow still helps, as in Figure.~\ref{fig:sbu} and Table.~\ref{table:detection}. 

\begin{table}[t]
\caption{Comparison on shadow detection.}
\label{table:detection}
\begin{center}
\begin{tabular}{|l|l|c|c|c|}
\hline
Method & backbone & S$\downarrow$ & NS$\downarrow$ & BER$\downarrow$  \\
\hline
\hline
scGAN$_{17'}$ & - &  8.39 & 9.69 & 9.04 \\
ST-CGAN$_{18'}$ & - & 3.75 & 12.53 & 8.14   \\
DSC$_{18'}$  &   VGG19    & 9.76  & \textbf{1.42} & 5.59   \\
ADNet$_{18'}$  & Adversarial &  4.45  &  6.30  & 5.37  \\
Ours(HAN)   &  VGG19 & 2.80 & 6.32 & 4.56 \\ 
Ours(HAN)+DA   &  VGG19 & \textbf{2.71} & 5.87 & 4.29 \\ 
\hline
BDRAR$_{18'}$  & \textbf{ResNeXt101} &  3.40 &  3.89  & 3.64 \\
DSDNet$_{19'}$  & \textbf{ResNeXt101} &  3.33  & 3.58    & \textbf{3.45} \\
\hline
\end{tabular}
\end{center}
\end{table}

\subsection{Ablation Studies}	
\subsubsection{Network Structure}
\begin{figure}[t]
 \centering
  \includegraphics[width=\columnwidth]{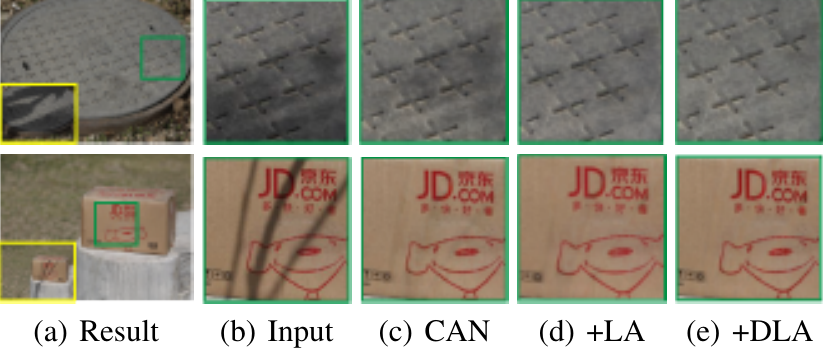}
  \caption{Ablation study on our network structure. We plot the results of our full methods with the input in yellow region. The baseline network CAN can remove the shadow but still have color in-consistency, and our feature LA(+LA) and dual LA(+DLA) can further remove the artifacts.}
  \label{fig:ablation}
\end{figure}

\begin{figure}[t]
 \centering
  \includegraphics[width=\columnwidth]{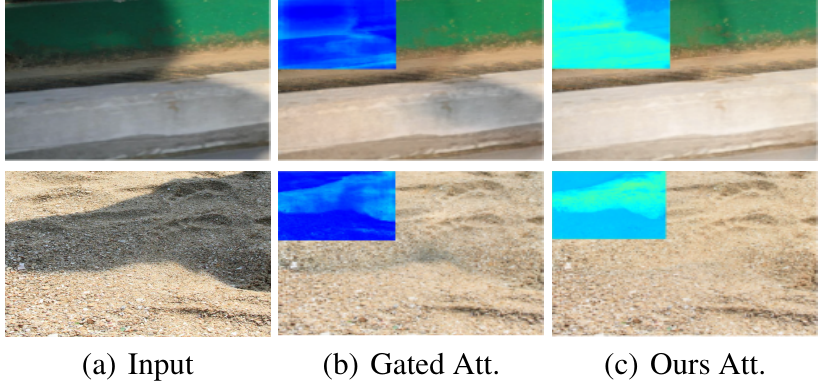}
  \caption{Comparison with the attention from gated convolution. We show the results and corresponding attention map. It is obviously that gated attention learns the feature locally while our attention predict the shadow region successfully.}
  \label{fig:gated}
\end{figure}

We evaluate the contributions of our network structure on the SRD dataset. As shown in Table.~\ref{table:eval} and Figure.~\ref{fig:ablation}, our baseline is the \textit{context aggregation network}~(CAN)~\cite{2018arXiv180605376Z} that is designed for image processing and reflection removal. It has a poor performance on the shadow removal task both on shadow and non-shadow regions. However, our feature layer aggregation~(+ LA) technique greatly improves the quality of the shadow regions. This is because the hierarchical mixed features will force the network to better understand multiple contexts. Then, when training using the dual aggregation network with attention aggregation~(+DLA), our method gains far better results with the supervision of the shadow mask. In addition, we find that the local gated convolution~\cite{yu2018free} is quantitatively less effective on our task. Here, we plot an example in Figure.~\ref{fig:gated} and compare our attention with the results on local gated convolution. 
We visualize the mean responses on the last attention features after $Sigmoid$. The figure shows that when the shadow region is large enough, the local gated convolution only gains the knowledge from its sibling convolution, causing the failure to capture the whole image context.

\begin{table}[b]
\caption{Ablation Study on SRD dataset.}
\begin{center}
\begin{tabular}{|l|c|c|c|}
\hline
Methods & S$\downarrow$ & NS$\downarrow$ & RMSE$\downarrow$  \\
\hline
\hline
CAN & 10.05 & 5.11 & 6.14 \\
+ LA  & 9.31 & 5.10 & 5.98 \\
+ DLA & \textbf{8.93} & \textbf{4.80} & \textbf{5.67} \\
\hline
\end{tabular}
\end{center}

\label{table:eval}
\end{table}

\subsubsection{Comparison on Shadow Synthesis}
Since there are few deep learning-based methods for shadow synthesis particularly, we compare the auxiliary network for unpaired shadow removal in the Mask-ShadowGAN~\cite{Anonymous:XfCvZjhb} and the ShadowGAN~\cite{zhang2019shadowgan}, which synthesizes shadows using the Pix2Pix~\cite{isola2017image}. Note that, the shadow generator in Mask-ShadowGAN is just an auxiliary helper for unpaired shadow removal, other than synthesizing the paired dataset for supervised learning as data augmentation. The ShadowGAN trains using only the low-resolution images from the computer rendered dataset. For comparison, we train our shadow synthesis network and others on the ISTD training set and test by synthesizing the shadow images using the ISTD test set. We evaluate the results using the SSIM and PSNR. In addition, since perceptual quality is also important for image generation, we evaluate the generated shadow using the image perceptual quality metrics LPIPS~\cite{zhang2018perceptual}. As shown in Table.~\ref{tab:shadow}, our method outperforms other methods in all the metrics.

\begin{table}[t]
\caption{Comparison on the quality of shadow synthesis.}
\begin{center}
\begin{tabular}{|l|c|c|c|}
\hline
Method & LPIPS $\downarrow$ & SSIM $\uparrow$ & PSNR $\uparrow$ \\
\hline
\hline
Masked-ShadowGAN & 25.30 & 80.18 & 23.18  \\
Pix2Pix & 11.51 & 88.82 & 26.01  \\
Ours & \textbf{5.38} & \textbf{91.67} & \textbf{26.31}  \\
\hline
\end{tabular}
\end{center}

\label{tab:shadow}
\end{table}

\subsubsection{Extend to Novel Domain}
\begin{figure}[t]
\centering     
\includegraphics[width=\columnwidth]{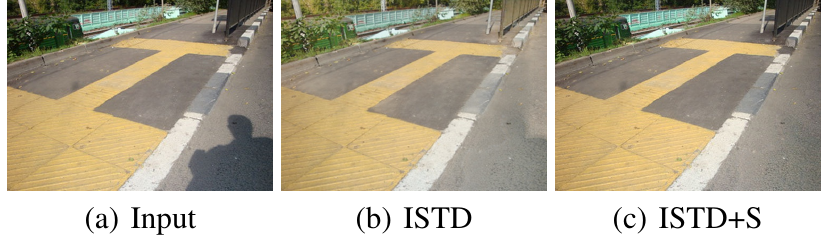}
\caption{Training using our synthesized dataset~(ISTD+S) helps the network works better on novel domains.}
\label{fig:novel}
\end{figure}

Training using our synthesized dataset enlarges the capabilities of the network. Here, we show the potential of our methods in novel scenes. Particularly, we test the pre-trained ISTD model on the SRD and SBU datasets \textbf{without} retraining on their specific training datasets. We show an examples from shadow detection datasets SBU in Figure.~\ref{fig:novel}. With the help of the synthesized dataset~(ISTD+S in Figure.~\ref{fig:novel}), our network can better understand the scenes and remove the shadows in these regions without ghosts or color inconsistencies. In the SRD dataset, when training using the ISTD+S dataset, the RMSE of our network is 9.82; however, if we train using only the ISTD dataset, the RMSE is 11.9. 

\section{Conclusion}
In this paper, we aim to create a ghost-free shadow removal method in two aspects. First, we propose a novel network that starts with context aggregation network and hierarchically aggregates the features and attentions. Our network structure can produce high-quality border-free results from multi-context features and attentions. Then, to enlarge the scenes in the current shadow removal dataset and reduce the color inconsistencies, we train a shadow matting GAN to synthesize the shadow images from the corresponding shadow-free images and masks. The proposed shadow generator is used to augment the data of the current dataset. Experiments show both our methods are helpful for generating the ghost-free shadow-free images. 

\section{Acknowledgments}
This work was partly supported by the University of Macau under Grants: MYRG2018-00035-FST and MYRG2019-00086-FST, the Science and Technology Development Fund, Macau SAR~(File no. 041-2017-A1) and the National Natural Science Foundation of China~(61902313).

\bibliographystyle{aaai}
\bibliography{shadow.bib}

\end{document}